\pdfoutput=1

\documentclass[11pt]{article}

\usepackage[]{ACL2023}

\usepackage{times}
\usepackage{latexsym}
\usepackage{bbm}
\usepackage{graphicx}
\usepackage{booktabs}
\usepackage{amssymb}
\usepackage{setspace}
\usepackage{subcaption}

\usepackage[ruled,vlined]{algorithm2e}
\RestyleAlgo{ruled}

\usepackage[T1]{fontenc}

\usepackage[utf8]{inputenc}

\usepackage{microtype}

\usepackage{inconsolata}

\title{Membership Inference Attacks against Language Models\\via Neighbourhood Comparison}

\author{Justus Mattern$^1$, Fatemehsadat Mireshghallah$^2$, Zhijing Jin$^{3,4}$,\\ \textbf{Bernhard Schölkopf}$^3$, \textbf{Mrinmaya Sachan}$^4$, \textbf{Taylor Berg-Kirkpatrick}$^2$ \\
 RWTH Aachen$^1$, UC San Diego$^2$, MPI for Intelligent Systems$^3$, ETH Zürich$^4$ \\
  \textit{Correspondence:} \texttt{justusmattern27@gmail.com}
 }

\begin{document}
\maketitle
\begin{abstract}

Membership Inference attacks (MIAs) aim to predict whether a data sample was present in the training data of a machine learning model or not, and are widely used for assessing the privacy risks of language models. Most existing attacks rely on the observation that models tend to
assign higher probabilities to their training samples than non-training points. However, simple thresholding of the model score in isolation tends to lead to high false-positive rates as it does not account for the intrinsic complexity of a sample. Recent work has demonstrated that reference-based attacks which compare model scores to those obtained from a reference model trained on similar 
data can substantially improve the performance of MIAs.
However, in order to train reference models, attacks of this kind make the strong and arguably unrealistic assumption that an adversary has access to samples closely resembling the original training data. Therefore, we investigate their performance in more realistic scenarios and find that they are highly fragile in relation to the data distribution used to train reference models. To investigate whether this fragility provides a layer of safety, we propose and evaluate neighbourhood attacks, which compare model scores for a given sample to scores of synthetically generated neighbour texts and therefore eliminate the need for access to the training data distribution. We show that, in addition to being competitive with reference-based attacks that have perfect knowledge about the training data distribution, our attack clearly outperforms existing reference-free attacks as well as reference-based attacks with imperfect knowledge, which demonstrates the need for a reevaluation of the threat model of adversarial attacks~\footnote{Code is available here: \url{https://github.com/mireshghallah/neighborhood-curvature-mia}}. 

\end{abstract}


\section{Introduction}

The public release and deployment of machine learning models trained on potentially sensitive user data introduces a variety of privacy risks: While embedding models have been shown to leak personal attributes of their data \citep{info-leakage-embedding}, generative language models are capable of generating verbatim repetitions of their training data and therefore exposing sensitive strings such as names, phone numbers or email-addresses \citep{lm-extractdata}.
Another source of risk arises from membership inference attacks (MIAs) \citep{Shokri2016MembershipIA}, which enable adversaries to classify whether a given data sample was present in a target model's training data or not. Due to their simplicity and the fact that MIAs are an important component of more sophisticated attacks such as extraction attacks \citep{lm-extractdata}, they have become one of the most widely used tools to evaluate data leakage and empirically study the privacy of machine learning models \citep{Murakonda2020MLPM, song2020introducing}.

\begin{figure*}[t]
    \centering
    \includegraphics[width=1.0\linewidth]{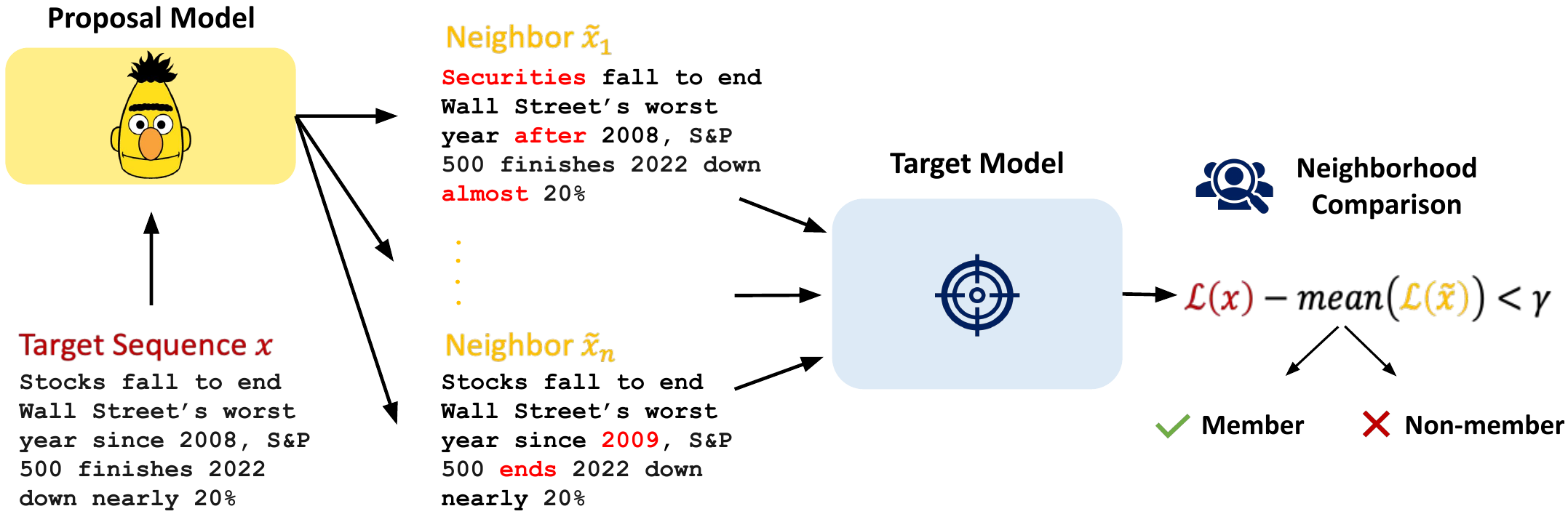}
    \caption{Overview of our attack: Given a target sample $x$, we use a pretrained masked language model to generate highly similar neighbour sentences through word replacements. Consequently, we compare our neighbours' losses and those of the original sample under the target model by computing their difference. As our neighbours are highly similar to the target sequence, we expect their losses to be approximately equal to the target model and only to be lower if the target sequence was a sample of the model's training data. In this case, the difference should be below our threshold value $\gamma$.}
\end{figure*}


Typically, membership inference attacks exploit models' tendency to overfit their training data and therefore exhibit lower loss values for training members \citep{loss-attack, Sablayrolles2019WhiteboxVB}. A highly simple and commonly used baseline attack is therefore the LOSS attack \citep{loss-attack}, which classifies samples as training members if their loss values are below a certain threshold.
While attacks of this kind do generally reap high accuracies, \citet{Carlini2021MembershipIA} point out a significant flaw: \textit{Good accuracies for attacks of this kind are primarily a result of their ability to identify non-members rather than training data members}, which does arguably not pose important privacy risks.
This shortcoming can be attributed to the fact that certain samples such as repetitive or very simple short sentences are naturally assigned higher probabilities than others \citep{fan-etal-2018-hierarchical, Holtzman2020The}, and the influence of this aspect on the obtained model score largely outweighs the influence of a model's tendency to overfit its training samples \citep{Carlini2021MembershipIA}. To account for this, previous work has introduced the idea of \textit{difficulty calibration mechanisms} \citep{Long2018UnderstandingMI, watson2022on}, which aim to quantify the intrinsic complexity of a data sample (i.e., how much of an outlier the given sample is under the probability distribution of the target model) and subsequently use this value to regularize model scores before comparing them to a threshold value.

In practice, difficulty calibration is mostly realized through \textit{Likelihood Ratio Attacks (LiRA)}, which measure the difficulty of a target point by feeding it to \textit{reference models} that help provide a perspective into how likely that target point is in the given domain \citep{enhanced-2022,Carlini2021MembershipIA, watson2022on, mlm-membership-inference, memorization-nlp-finetuning}.
In order to train such reference models, LiRAs assume that an adversary has knowledge about the distribution of the target model's training data and access to a sufficient amount of samples from it. We argue that this is a highly optimistic and in many cases unrealistic assumption: as also pointed out by \citet{tramer-considerations}, in applications in which we care about privacy and protecting our models from leaking data (e.g. in the medical domain), high-quality, public in-domain data may not be available, which renders reference-based attacks ineffective. Therefore, we aim to design an attack which does not require any additional data:
For the design of our proposed \textit{neighborhood attack}, we build on the intuition of using references to help us infer membership, but instead of using reference models, we use \textit{neighboring samples}, which are textual samples crafted through data augmentations such as word replacements to be  
non-training members that are as similar as possible to the target point and therefore practically interchangeable with it in almost any context. With the intuition that neighbors should be assigned equal probabilities as the original sample under any plausible textual probability distribution,
we then compare the model scores of all these neighboring points to that of the target point and classify its membership based on their difference. Similar to LiRAs, we hypothesize that if the model score of the target data is similar to the crafted neighbors, then they are all plausible points from the distribution and the target point is not a member of the training set. However, if a sample is much more likely under the target model's distribution than its neighbors, we infer that this could only be a result of overfitting and therefore the sample must be a part of the model's training data.

We conduct extensive experiments measuring the performance of our proposed neighborhood attack, and particularly compare it to reference-based attacks with various different assumptions about knowledge of the target distribution and access to additional data. Concretely, amongst other experiments, we simulate real-world reference-based attacks by training reference models on external datasets from the same domain as the target model's training data. We find
that neighbourhood attacks outperform LiRAs with more realistic assumptions about the quality of accessible data by up to 100\%, and even show competitive performance when we assume that an attacker has perfect knowledge about the target distribution and access to a large amount of high-quality samples from it.

\section{Membership Inference Attacks via Neighbourhood Comparison}

In this section, we provide a detailed description of our attack, starting with the general idea of comparing neighbouring samples and following with a technical description of how to generate such neighbors.

\subsection{General Idea}

We follow the commonly used setup of membership inference attacks in which the adversary has grey-box access to a machine learning model $f_\theta$ trained on an unknown dataset $\mathcal{D}_{\mathrm{train}}$, meaning that they can obtain confidence scores and therefore loss values from $f_\theta$, but no additional information such as model weights or gradients. The adversary's goal is to learn an attack function $A_{f_\theta} : \mathcal{X} \rightarrow \{0, 1\}$ , which determines for each $x$ from the universe of textual samples $\mathcal{X}$ whether $x \in D_{\mathrm{train}}$ or $x \not\in D_{\mathrm{train}}$. As mentioned in the previous section, the LOSS attack \citep{loss-attack}, one of the most simple forms of membership inference attacks, classifies samples by thresholding their loss scores, so that the membership decision rule is:

\begin{equation}
\label{eq:lossattack}
    A_{f_\theta}(x) = \mathbbm{1}[\mathcal{L}(f_\theta, x) < \gamma].
\end{equation}

More recent attacks follow a similar setup, but perform difficulty calibration to additionally account for the intrinsic complexity of the sample $x$ under the target distribution and adjust its loss value accordingly. Concretely, given a function $d: \mathcal{X} \rightarrow \mathbb{R}$ assigning difficulty scores to data samples, we can extend the 
the decision rule to

\begin{equation}
\label{eq:loss}
    A_{f_\theta}(x) = \mathbbm{1}[\mathcal{L}(f_\theta, x) - d(x) < \gamma].
\end{equation}

Likelihood Ratio Attacks (LiRAs)~\cite{enhanced-2022}, the currently most widely used form of membership inference attacks, use a sample's loss score obtained from some reference model $f_\phi$ as a difficulty score, so that $d(x) = \mathcal{L}(f_\phi, x)$ . However, this makes the suitability of the difficulty score function dependent on the quality of reference models and therefore the access to data from the training distribution. We circumvent this by designing a different difficulty calibration function depending on synthetically crafted neighbors.

Formally, for a given $x$, we aim to produce natural adjacent samples, or a set of $n$ neighbors $\{\tilde{x}_1, ..., \tilde{x}_n\}$, which slightly differ from $x$ and are not part of the target model's training data, but are approximately equally likely to appear in the general distribution of textual data, and therefore offer a meaningful comparison. Given our set of neighbors, we calibrate the loss score of $x$ under the target model by subtracting the average loss of its neighbors from it, resulting in a new decision rule:



\begin{equation}
\label{eq:neighbourattack}
    A_{f_\theta}(x) = \mathbbm{1}\left[\left(\mathcal{L}(f_\theta, x) - \sum_{i=1}^n \frac{\mathcal{L}(f_\theta, \tilde{x}_i)}{n}\right) < \gamma\right].
\end{equation}

The interpretation of this decision rule is straightforward: 
Neighbors crafted through minimal changes that fully preserve the semantics and grammar of a given sample should in theory be interchangeable with the original sentence and therefore be assigned highly similar likelihoods under any textual probability distribution.
Assuming that our neighbors were not present in the training data of the target model, we can therefore use the model score assigned to them as a proxy for what the original sample's loss should be if it was not present in the training data.
The target sample's loss value being substantially lower than the neighbors' losses could therefore only be a result of overfitting and therefore the target sample being a training member. In this case, we expect the difference in Equation \ref{eq:neighbourattack} to be below our threshold value $\gamma$


\subsection{Obtaining Neighbour Samples}

In the previous section, for a given text $x$, we assumed  access to a set of adjacent samples $\{\tilde{x}_1, ..., \tilde{x}_n\}$. In this section we describe how those samples are generated. As it is highly important to consider neighbours that are approximately equally complex, it is important to mention that beyond the semantics of $x$, we should also preserve structure and syntax, and can therefore not simply consider standard textual style transfer or paraphrasing models. Instead, we opt for very simple word replacements that preserve semantics and fit the context of the original word well. For obtaining these replacements, we adopt the framework proposed by \citet{zhou-etal-2019-bert}, who propose the use of transformer-based \citep{NIPS2017_3f5ee243} masked language models (MLMs) such as BERT \citep{devlin-etal-2019-bert} for lexical substitutions: Concretely, given a text $x := (w^{(1)}, ..., w^{(L)})$ consisting of $L$ tokens, the probability $p_\theta(\tilde{w} = w^{(i)} | x)$ of token $\tilde{w}$ as the word in position $i$ can be obtained from the MLM's probability distribution $p(\mathcal{V}^{(i)}|x)$ over our token vocabulary $\mathcal{V}$ at position $i$. As we do not want to consider the influence of the probability of the original token on the token's suitability as a replacement when comparing it to other candidates, we normalize the probability over all probabilities except that of the original token. So, if $\hat{w}$ was the original token at position $i$, our suitability score for $\tilde{w}$ as a replacement is

\begin{equation}
    p_{\mathrm{swap}}(\hat{w}^{(i)}, \tilde{w}^{(i)}) = \frac{p_\theta(\tilde{w} = w^{(i)} | x)}{1 - p_\theta(\hat{w} = w^{(i)} | x)}.
    \label{eq:suitability}
\end{equation}

In practice, simply masking the token which we want to replace will lead to our model completely neglecting the meaning of the original word when predicting alternative tokens and therefore potentially change the semantics of the original sentence -- for instance, for the given sample "The movie was great", the probability distribution for the last token obtained from "The movie was [MASK]" might assign high scores to negative words such as "bad", which are clearly not semantically suitable replacements. To counteract this, \citet{zhou-etal-2019-bert} propose to keep the original token in the input text, but to add strong dropout to the input embedding layer at position $i$ before feeding it into the transformer to obtain replacement candidates for $w^{(i)}$. We adopt this technique, and therefore obtain a procedure which allows us to obtain $n$ suitable neighbors with $m$ word replacements using merely an off-the-shelf model that does not require any adaptation to the target domain. The pseudocode is outlined in Algorithm~\ref{alg:algo}.


\begin{algorithm}[h]
  \microtypesetup{protrusion=false} 

  \SetKwInOut{KwIn}{Input}
  \SetKwInOut{KwOut}{Output}
  \KwIn{Text $x = (w^{(1)}, ..., w^{(L)})$, $n$, $m$}
  \KwOut{Neighbours $\{\tilde{x}_1, ..., \tilde{x}_n\}$ with $m$ word replacements each}
\begin{spacing}{1.2}
  \For{\(i \in \{1, \ldots, L\}\)}
  {
   Get embeddings $(\phi(w^{(1)}), ..,\phi(w^{(L)})$.\\
   Add dropout: $\phi(w^{(i)}) = \mathrm{drop}(\phi(w^{(i)}))$.\\
   Obtain $p(\mathcal{V}^{(i)}|x)$ from BERT.\\
   Compute $p_{\mathrm{swap}}(w^{(i)}, \tilde{w}^{(i)}) \forall \tilde{w} \in \mathcal{V}$. 
  }
  \end{spacing}
  For all swaps $(w^{(i_1)}, \tilde{w}^{(i_1)}) ... (w^{(i_m)}, \tilde{w}^{(i_m)})$ with $i_k \neq i_l$ for $i \neq l$, compute joint suitability $\sum_{i=1}^m p_{\mathrm{swap}}(w^{(i_1)}, \tilde{w}^{(i_1)})$ and return $n$ highest

  \caption{Neighbourhood Generation}
  \label{alg:algo}

\end{algorithm}

\section{Experimental Setup}
\label{sec:experiments}

We evaluate the performance of our attack as well as reference-free and reference-based baseline attacks against large autoregressive models trained with the classical language modeling objective. Particularly, we use the base version of GPT-2 \citep{radford2019language} as our target model.

\subsection{Datasets}

We perform experiments on three datasets, particularly news article summaries obtained from a subset of the AG News corpus\footnote{\url{http://groups.di.unipi.it/~gulli/AG_corpus_of_news_articles.html}} containing four news categories ("World", "Sports", "Business", "Science \& Technology"), tweets from the Sentiment140 dataset \citep{go2009twitter} and excerpts from wikipedia articles from Wikitext-103 \citep{merity2017pointer}. Both datasets are divided into two disjunct subsets of equal size: one of these subsets serves as training data for the target model and therefore consists of positive examples for the membership classification task. Subset two is not used for training, but its samples are used as negative examples for the classification task. The subsets contain 60,000, 150,000 and 100,000 samples for AG News, Twitter and Wikitext, respectively, leading to a total size of 120,000, 300,000 and 200,000 samples. For all corpora, we also keep an additional third subset that we can use to train reference models for reference-based attacks.

\subsection{Baselines}
To compare the performance of our attack, we consider various baselines: As the standard method for reference-free attacks, we choose the \textbf{LOSS Attack} proposed by \citet{loss-attack}, which classifies samples as training members or non-members based on whether their loss is above or below a certain threshold (see Equation \ref{eq:lossattack}). For reference-based attacks, we follow recent implementations \citep{mlm-membership-inference, memorization-nlp-finetuning, watson2022on} and use reference data to train a single reference model of the same architecture as the target model. Subsequently, we measure whether the likelihood of a sample under the target model divided by its likelihood under the reference model crosses a certain threshold.

\paragraph{Training Data for Reference Models}
As discussed in previous sections, we would like to evaluate reference-based attacks with more realistic assumptions about access to the training data distribution. Therefore, we use multiple reference models trained on different datasets: As our \textbf{Base Reference Model}, we consider the pretrained, but not fine-tuned version of GPT-2. Given the large pretraining corpus of this model, it should serve as a good estimator of the general complexity of textual samples and has also been successfully used for previous implementations of reference-based attacks \citep{memorization-nlp-finetuning}. Similar to our neighbourhood attack, this reference model does not require an attacker to have any additional data or knowledge about the training data distribution.

To train more powerful, but still realistic reference models, which we henceforth refer to as \textbf{Candidate Reference Models}, we use data that is in general similar to the target model's training data, but slightly deviates with regard to topics or artifacts that are the result of the data collection procedure.
Concretely, we perform this experiment for both our AG News and Twitter corpora: For the former, we use article summaries from remaining news categories present in the AG News corpus ("U.S.", "Europe", "Music Feeds", "Health", "Software and Development", "Entertainment") as well as the NewsCatcher dataset\footnote{\url{https://github.com/kotartemiy/topic-labeled-news-dataset}} containing article summaries for eight categories that highly overlap with AG News ("Business", "Entertainment", "Health", "Nation", "Science", "Sports", "Technology", "World"). For Twitter, we use a depression detection dataset for mental health support from tweets \footnote{\url{https://www.kaggle.com/datasets/infamouscoder/mental-health-social-media}} as well as tweet data annotated for offensive language \footnote{\url{https://www.kaggle.com/datasets/mrmorj/hate-speech-and-offensive-language-dataset}}. As it was highly difficult to find data for reference models, it was not always possible to match the amount of training samples of the target model. The number of samples present in each dataset can be found in Table \ref{tab:numsamples}.

\begin{table}[t]
\small
  \centering
\setlength\tabcolsep{10pt}
  \begin{tabular}{ l  c }
    \toprule
    Dataset & \#Samples\\
\midrule
AG News (Other Categories) & 60,000\\
NewsCatcher & 60,000\\
AG News Oracle Data & 60,000\\
\midrule
Twitter Mental Health & 20,000\\
Twitter Offensive Language & 25,000\\
Twitter Oracle Data & 150,000\\
\midrule
Wikipedia Oracle Data & 100,000\\
\bottomrule
  \end{tabular}
  \caption{Number of samples in the reference model training data. Target models for News, Twitter and Wikipedia were trained on 60,000, 150,000 and 100,000 samples, respectively.}
  \label{tab:numsamples}
\end{table}

As our most powerful reference model, henceforth referred to as \textbf{Oracle Reference Model}, we use models trained on the same corpora, but different subsets as the target models. This setup assumes that an attacker has perfect knowledge about the training data distribution of the target model and high quality samples.

\begin{table*}[t]
    \small
  \centering

\setlength\tabcolsep{6pt}

  \begin{tabular}{ l  c c c c c c c c c}
    \toprule
    & \multicolumn{3}{c}{News}& \multicolumn{3}{c}{Twitter} & \multicolumn{3}{c}{Wikipedia}\\
    \cmidrule(lr){2-4} \cmidrule(lr){5-7} \cmidrule(lr){8-10}
    False Positive Rate & 1\% & 0.1\% &  0.01\% & 1\% & 0.1\% & 0.01\% & 1\% & 0.1\% & 0.01\%\\
    \midrule
    \textbf{Likelihood Ratio Attacks:}&&&&&&&&&\\
    Base Reference Model & 4.24\% & 0.91\% & 0.16\% & 5.66\% & 0.98\% & 0.22\% & 1.21\% & 0.12\% & 0.01\%\\
    Candidate Reference Model 1 & 4.91\% & 0.95\% & 0.15\% & 6.49\% & 1.10\% & 0.24\% & &&\\
    Candidate Reference Model 2 & 4.76\% & 0.92\% & 0.15\% & 6.61\% &1.19\% & 0.25\% & &&\\
    Oracle Reference Model* & 18.90\% & 3.76\% & 0.16\% & 13.90\% & 1.59\% & 0.28\% & 11.70\% & 3.70\% & 0.12\%\\
    \midrule
    \textbf{Reference-Free Attacks:}&&&&&&&&&\\
    LOSS Attack & 3.50\% & 0.10\% & 0.01\% & 2.08\% & 0.11\% & 0.02\% & 1.06\% & 0.11\% & 0.01\%\\
    Neighbour Attack (Ours) & \textbf{8.29\%} & \textbf{1.73\%} & \textbf{0.29\%} & \textbf{7.35\%} & \textbf{1.43\%} & \textbf{0.28\%} & \textbf{2.32\%} & \textbf{0.27\%} & \textbf{0.10\%}\\
    \bottomrule

  \end{tabular}

  \caption{
  True positive rates of various attacks for low false positive rates of $1\%, 0.1\%$, and $0.01\%$. Candidate Reference Model 1 refers to reference models trained on data from other AG News categories and our Twitter mental health dataset, Candidate Reference Model 2 refers to reference models trained on NewsCatcher and the offensive tweet classification dataset. *As reference attacks trained on oracle datasets represent a rather unrealistic scenario with perfect assumptions, we compare our results with other baselines with more realistic assumptions when highlighting best results as bold.}
  \label{tab:results}
\end{table*}



\subsection{Implementation Details}
We obtain and fine-tune all pretrained models using the Huggingface transformers library \citep{wolf-etal-2020-transformers} and PyTorch \citep{NEURIPS2019_bdbca288}. 
As target models, we fine-tune the pretrained 117M parameter version of GPT-2, which originally has a validation perplexity of 56.8 and 200.3 on AG News and Twitter data, respectively, up to validation set perplexities of 30.0 and 84.7. In our initial implementation of our neighbourhood attack, we obtain the 100 most likely neighbour samples using one word replacement only from the pretrained 110M parameter version of BERT. We apply a dropout of $p=0.7$ to the embedding of the token we want to replace.
For evaluating LiRA baselines, we train each reference model on its respective training dataset over multiple epochs, and choose the best performing reference model w.r.t attack performance. Following \citet{Carlini2021MembershipIA}, we evaluate our attack's precision for predetermined low false positive rate values such as 1\% or 0.01\%. We implement this evaluation scheme by adjusting our threshold $\gamma$ to meet this requirement and subsequently measure the attack's precision for the corresponding $\gamma$.
All models have been deployed on single GeForce RTX 2080 and Tesla K40 GPUs.

\section{Results}

In this section, we report our main results and perform additional experiments investigating the impact of reference model performance on the success of reference-based attacks as well as several ablation studies.
Following \citep{Carlini2021MembershipIA}, we report attack performances in terms of their true positive rates (TPR) under very low false positive rates (FPR) by adjusting the threshold value $\gamma$. Concretely, we choose 1\%, 0.1\% and 0.01\% as our target FPR values.


\subsection{Main Results}
Our results can be found in Table \ref{tab:results} and \ref{tab:aucresults}, with the former showing our attack performance in terms of true positive rates under low false positive rates and the latter showing AUC values. As previously discovered, the LOSS attack tends to perform badly when evaluated for very low false positive rates \citep{Carlini2021MembershipIA, watson2022on}. Likelihood Ratio Attacks can clearly outperform it, but we observe that their success is highly dependent on having access to suitable training data for reference models: Attacks using the base reference models and candidate models can not reach the performance of an attack using the oracle reference model by a large margin. Notably, they are also substantially outperformed by our Neighbour Attack, which can, particularly in low FPR ranges, even compete very well with or outperform Likelihood Ratio Attacks with an Oracle Reference Model, without relying on access to any additional data.

\begin{table}[h]
    \small
  \centering

\setlength\tabcolsep{6pt}

  \begin{tabular}{ l  c c c}
    \toprule
    & News & Twitter & Wiki \\
    \midrule
    \textbf{LiRA:}&&&\\
    Base Reference Model & 0.76 & 0.75 & 0.54\\
    Candidate Reference 1 & 0.78 & \textbf{0.81} &  \\
    Candidate Reference 2 & 0.75 & 0.77 &\\
    Oracle Reference* & 0.94 & 0.89 & 0.90 \\
    \midrule
    \textbf{Other Attacks:}&&&\\
    LOSS Attack & 0.64 & 0.60 & 0.52\\
    Neighbour Attack & \textbf{0.79} & 0.77 & \textbf{0.62}\\
    \bottomrule

  \end{tabular}
  \caption{
  AUC values of various attacks.}
  \label{tab:aucresults}
\end{table}

\subsection{Measuring the Dependence of Attack Success on Reference Model Quality}

Motivated by the comparably poor performance of Likelihood Ratio Attacks with reference models trained on only slightly different datasets to the target training data, we aim to investigate the dependence of reference attack performances on the quality of reference models in a more controlled and systematic way. To do so, we train reference models on our oracle data over multiple epochs, and report the attack performance of Likelihood Ratio Attacks w.r.t to the reference models' validation perplexity (PPL) on a held out test set, which is in this case the set of non-training members of the target model. Intuitively, we would expect the attack performance to peak when the validation PPL of reference models is similar to that of the target model, as this way, the models capture a very similar distribution and therefore offer the best comparison to the attack model. In this setup, we are however particularly interested in the attack performance when the validation PPL does not exactly match that of the target model, given that attackers will not always be able to train perfectly performing reference models.

\begin{figure}[t]
\centering
\begin{subfigure}{\linewidth}
   \includegraphics[width=1\linewidth]{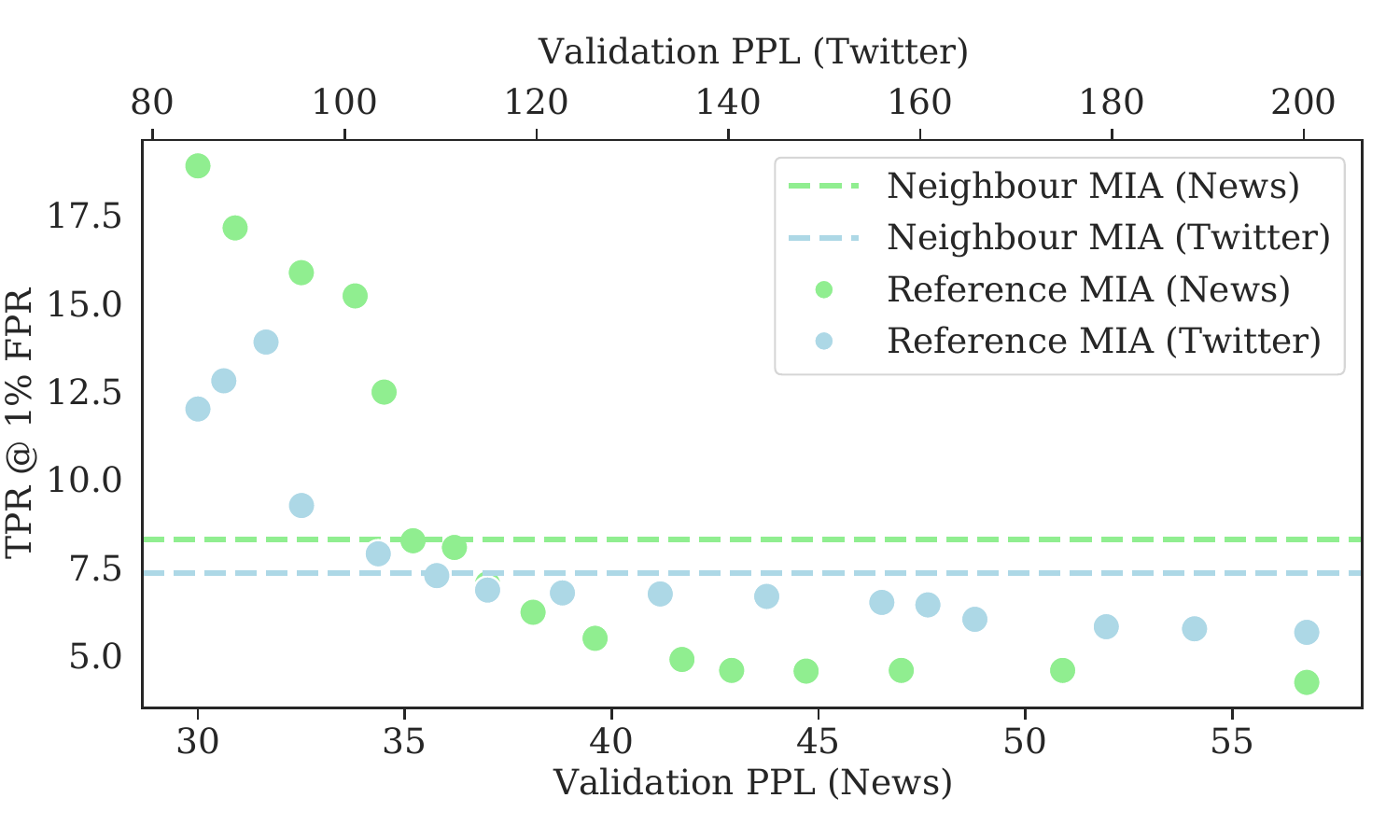}
\end{subfigure}
\begin{subfigure}{\linewidth}
   \includegraphics[width=1\linewidth]{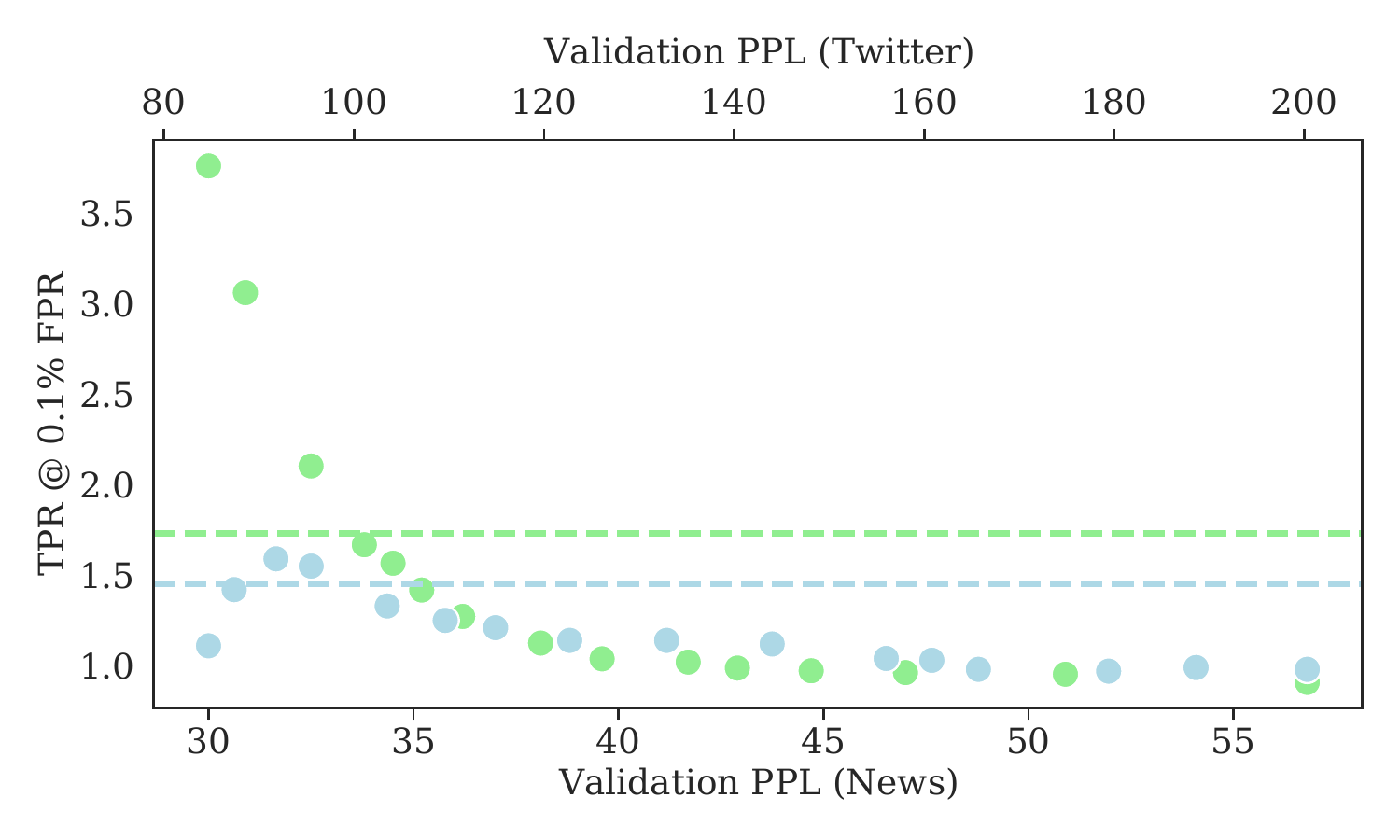}
\end{subfigure}
\begin{subfigure}{\linewidth}
   \includegraphics[width=1\linewidth]{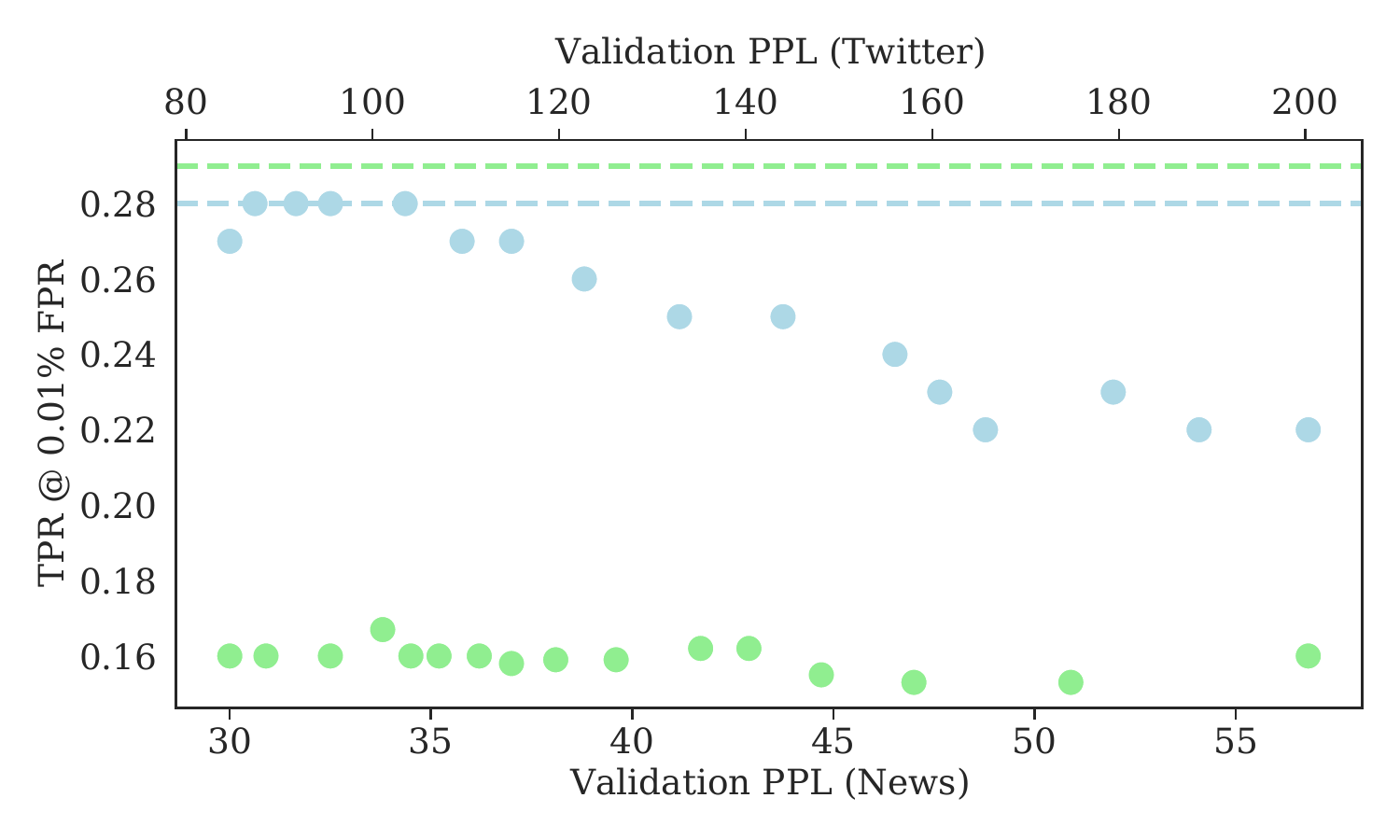}
\end{subfigure}
\caption{Attack Performance of reference attacks w.r.t validation PPL of reference models, compared to the performance of neighborhood attacks. The perplexities of the target models were 30.0 and 84.7 for AG News and Twitter, respectively}
\label{fig:epochstudy}

\end{figure}

The results of this experiment can be found in Figure \ref{fig:epochstudy} for our News and Twitter dataset and in Figure \ref{fig:epochstudywiki} for Wikitext. As can be seen, the performance of reference-based attacks does indeed peak when reference models perform roughly the same as the target model. A further very interesting observation is that substantial increases in attack success only seem to emerge as the validation PPL of reference models comes very close to that of the target model and therefore only crosses the success rate of neighbourhood attacks when the reference model's performance is almost the same as that of the target model. This further illustrates the fragility of reference-based attacks with respect to the choice of the reference model.

\subsection{Ablation Studies}
Having extensively studied the impact of different reference model training setups for the Likelihood Ratio Attack, we now aim to explore the effect of various components of our proposed neighbourhood attack.

\paragraph{Number of Generated Neighbours} For our main results in Table \ref{tab:results}, we report the performance of neighbour attacks for the 100 most likely generated neighbours as determined by BERT. In the following, we measure how varying this number affects the attack performance. While intuitively, a higher number of neighbours might offer a more robust comparison, it is also plausible that selecting a lower number of most likely neighbours under BERT will lead to neighbours of higher quality and therefore a more meaningful comparison of loss values. Our results in Table \ref{tab:numneighbours} show a clear trend towards the former hypothesis: The number of neighbours does in general have a strong influence on the performance of neighbourhood attacks and higher numbers of neighbours produce better results.

\begin{table}[t]
    \small
    \centering

\setlength\tabcolsep{4.5pt}
  \begin{tabular}{ l  c c c c c}

    \toprule
    \#Neighbours & 5 & 10 & 25 & 50 & 100\\
    \midrule
    \textbf{News:}&&&&&\\
    1\% FPR & 2.98\% & 4.57\% & 6.65\% & 8.19\% & 8.29\%\\
    0.1\% FPR & 0.53\% & 0.79\% & 1.43\% & 1.50\% & 1.73\%\\
    0.01\% FPR & 0.05\% & 0.07\% & 0.18\% & 0.23\% & 0.29\% \\
    \midrule
    \textbf{Twitter:}&&&&&\\
    1\% FPR & 3.93\% & 4.88\% & 6.21\% & 6.63\% & 7.35\%\\
    0.1\% FPR & 0.57\% & 0.62\% & 1.01\% & 1.34\% & 1.43\%\\
    0.01\% FPR & 0.05\% & 0.07\% & 0.10\% & 0.23\% & 0.28\%\\
    \midrule
    \textbf{Wikipedia:}&&&&&\\
    1\% FPR & 1.57\% & 1.81\% & 2.02\% & 2.17\% & 2.32\%\\
    0.1\% FPR & 0.16\% & 0.21\% & 0.23\% & 0.26\% & 0.27\%\\
    0.01\% FPR & 0.05\% & 0.08\% & 0.09\% & 0.10\% & 0.10\%\\
    \bottomrule

  \end{tabular}
  \caption{Attack performance w.r.t the number of neighbours against which we compare the target sample
  }
  \label{tab:numneighbours}
\end{table}

\paragraph{Number of Word Replacements}
Besides the number of generated neighbours, we study how the number of replaced words affects the performance of our attack.
While we reported results for the replacement of a single word in our main results in Table \ref{tab:results}, there are also reasons to expect that a higher number of replacements leads to better attack performance: While keeping neighbours as similar to the original samples as possible ensures that their probability in the general distribution of textual data remains as close as possible, one could also expect that too few changes lead the target model to assign the original sample and its neighbours almost exactly the same score, and therefore make it hard to observe high differences in loss scores for training members.
Our results of generating 100 neighbours with multiple word replacements are reported in Table \ref{tab:numreplacements}. We find that replacing only one word clearly outperforms multiple replacements. Beyond this, we do not find highly meaningful differences between two and three word replacements.

\begin{table}[t]
    \small
    \centering

\setlength\tabcolsep{6pt}
  \begin{tabular}{ l  c c c}

    \toprule
    \#Word Replacements & 1 & 2 & 3\\
    \midrule
    \textbf{News:}&&&\\
    1\% FPR & 8.29\% & 4.09\% & 4.18\%\\
    0.1\% FPR & 1.73\% & 0.85\% & 0.94\%\\
    0.01\% FPR & 0.29\%  & 0.23\% & 0.21\% \\
    \midrule
    \textbf{Twitter:}&&&\\
    1\% FPR & 7.35\% & 4.86\% & 4.37\%\\
    0.1\% FPR & 1.43\% & 0.74\% & 0.72\%\\
    0.01\% FPR & 0.28\% & 0.14\% & 0.11\% \\
    \midrule
    \textbf{Wikipedia:}&&&\\
    1\% FPR & 2.32\% & 1.76\% & 1.44\%\\
    0.1\% FPR & 0.27\% & 0.23\% & 0.17\%\\
    0.01\% FPR & 0.10\% & 0.07\% & 0.03\%\\
    \bottomrule

  \end{tabular}
  \caption{Attack performance w.r.t the number of words that are replaced when generating neighbours}
  \label{tab:numreplacements}
\end{table}

\section{Defending against Neighbourhood Attacks}

Due to the privacy risks that emerge from the possibility of membership inference and data extraction attacks, the research community is actively working on defenses to protect models.
Beyond approaches such as confidence score perturbation \citep{Jia2019MemGuardDA} and specific regularization techniques \citep{mireshghallah-etal-2021-privacy, chen2022relaxloss} showing good empirical performance, differentially private model training is one of the most well known defense techniques offering mathematical privacy guarantees: DP-SGD \citep{dpsgd-song, dpsgd-bassily, dpsgd}, which uses differential privacy \citep{dp-dwork} to bound the influence that a single training sample can have on the resulting model and has been shown to successfully protect models against membership inference attacks \citep{Carlini2021MembershipIA} and has recently also successfully been applied to training language models \citep{yu2022differentially, li2022large,mireshghallahdifferentially}. To test the effectiveness of differential privacy as a defense against neighbourhood attacks, we follow \citet{li2022large} and train our target model GPT-2 in a differentially private manner on AG News, where our attack performed the best. The results can be seen in Table \ref{tab:dpresults} and clearly demonstrate the effectiveness of DP-SGD. Even for comparably high epsilon values such as ten, the performance of the neighbourhood attack is substantially worse compared to the non-private model and is almost akin to random guessing for low FPR values.

\begin{table}[h]
    \small
    \centering

\setlength\tabcolsep{6pt}
  \begin{tabular}{ l  c c c c c}

    \toprule
    \ & $\epsilon=5$ & $\epsilon=10$ & $\epsilon=\infty$ \\
    \midrule
    TPR @ 1\% FPR & 1.29\% & 1.52\% & 8.29\% \\
    TPR @ 0.1\% FPR & 0.09\% & 0.13\% & 1.73\% \\
    TPR @ 0.01\% FPR & 0.01\% & 0.01\% & 0.29\% \\

    \bottomrule

  \end{tabular}
  \caption{Performance of neighbourhood attacks against models trained with DP-SGD}
  \label{tab:dpresults}
\end{table}


\section{Related Work}
\label{sec:relatedwork}
MIAs have first been proposed by \citet{Shokri2016MembershipIA} and continue to remain a topic of interest for the machine learning community. While many attacks, such as ours, assume to only have access to model confidence or loss scores \citep{loss-attack, Sablayrolles2019WhiteboxVB, Jayaraman2020RevisitingMI, watson2022on}, others exploit additional information such as model parameters \citep{stolen-memories} or training loss trajectories \citep{loss-trajectory-meminference}. Finally, some researchers have also attempted to perform membership inference attacks given only hard labels without confidence scores \citep{label-only-attack, pmlr-v139-choquette-choo21a}. Notably, the attack proposed by \citet{pmlr-v139-choquette-choo21a} is probably closest to our work as it tries to obtain information about a sample's membership by flipping its predicted labels through small data augmentations such as rotations to image data. To the best of our knowledge, we are the first to apply data augmentations of this kind for text-based attacks.

\paragraph{Membership Inference Attacks in NLP}
Specifically in NLP, membership inference attacks are an important component of language model extraction attacks \citep{lm-extractdata, memorization-nlp-finetuning}. Further studies of interest include work by \citet{hisamoto-etal-2020-membership}, which studies membership inference attacks in machine translation, as well as work by \citet{mlm-membership-inference}, which investigates Likelihood Ratio Attacks for masked language models. Specifically for language models, a large body of work also studies the related phenomenon of memorization \citep{Kandpal2022Deduplicating, carlini2022the, quantifying-memorization, counterfactual-memorization}, which enables membership inference and data extraction attacks in the first place.

\paragraph{Machine-Generated Text Detection} Due to the increasing use of tools like ChatGPT as writing assistants, the field of machine-generated text detection has become of high interest within the research community and is being studied extensively \citep{chakraborty2023possibilities, krishna2023paraphrasing, mitchell2023detectgpt,mireshghallah2023smaller}. Notably, \citet{mitchell2023detectgpt} propose DetectGPT, which works similarly to our attack as it compares the likelihood of a given sample under the target model to the likelihood of perturbed samples and hypothesizes that the likelihood of perturbations is smaller than that of texts the model has generated itself.

\section{Conclusion and Future Work}
In this paper, we have made two key contributions: First, we thoroughly investigated the assumption of access to in-domain data for reference-based membership inference attacks: In our experiments, we have found that likelihood ratio attacks, the most common form of reference-based attacks, are highly fragile to the quality of their reference models and therefore require attackers to have access to high-quality training data for those. Given that specifically in privacy-sensitive settings where publicly available data is scarce, this is not always a realistic assumption, we proposed that the design of reference-free attacks would simulate the behavior of attackers more accurately. Thus, we introduced neighborhood attacks, which calibrate the loss scores of a target samples using loss scores of plausible neighboring textual samples generated through word replacements, and therefore eliminate the need for reference trained on in-domain data. We have found that under realistic assumptions about an attacker's access to training data, our attack consistently outperforms reference-based attacks. Furthermore, when an attacker has perfect knowledge about the training data, our attack still shows competitive performance with reference-based attacks. We hereby further demonstrated the privacy risks associated with the deployment of language models and therefore the need for effective defense mechanisms. Future work could extend our attack to other modalities, such as visual or audio data, or explore our attack to improve extraction attacks against language models.

\section*{Limitations}

\paragraph{The proposed attack is specific to textual data}

While many membership inference attacks are universally applicable to all modalities as they mainly rely on loss values obtained from models, our proposed method for generating neighbours is specific to textual data. While standard augmentations such as rotations could be used to apply our method for visual data, this is not straightforward such as the transfer of other attacks to different modalities.

\paragraph{Implementation of baseline attacks}

As the performance of membership inference attacks depend on the training procedure of the attacked model as well as its degree of overfitting, it is not possible to simply compare attack performance metrics from other papers to ours. Instead, we had to reimplement existing attacks to compare them to our approach. While we followed the authors' descriptions in their papers as closely as possible, we cannot guarantee that their attacks were perfectly implemented and the comparison to our method is therefore 100\% fair.

\section*{Ethical Considerations}

Membership inference attacks can be used by malicious actors to compromise the privacy of individuals whose data has been used to train models. However, studying and expanding our knowledge of such attacks is crucial in order to build a better understanding for threat models and to build better defense mechanisms that take into account the tools available to malicious actors. Due to the importance of this aspect, we have extensively highlighted existing work studying how to defend against MIAs in Section \ref{sec:relatedwork}.
As we are aware of the potential risks that arise from membership inference attacks, we will not freely publicize our code, but instead give access for research projects upon request.
\newline\newline
With regards to the data we used, we do not see any issues as all datasets are publicly available and have been used for a long time in NLP research or data science competitons.

%
%
%



\bibliography{anthology,custom}
\bibliographystyle{acl_natbib}

\appendix

\begin{figure}[t]
\centering
\begin{subfigure}{\linewidth}
   \includegraphics[width=1\linewidth]{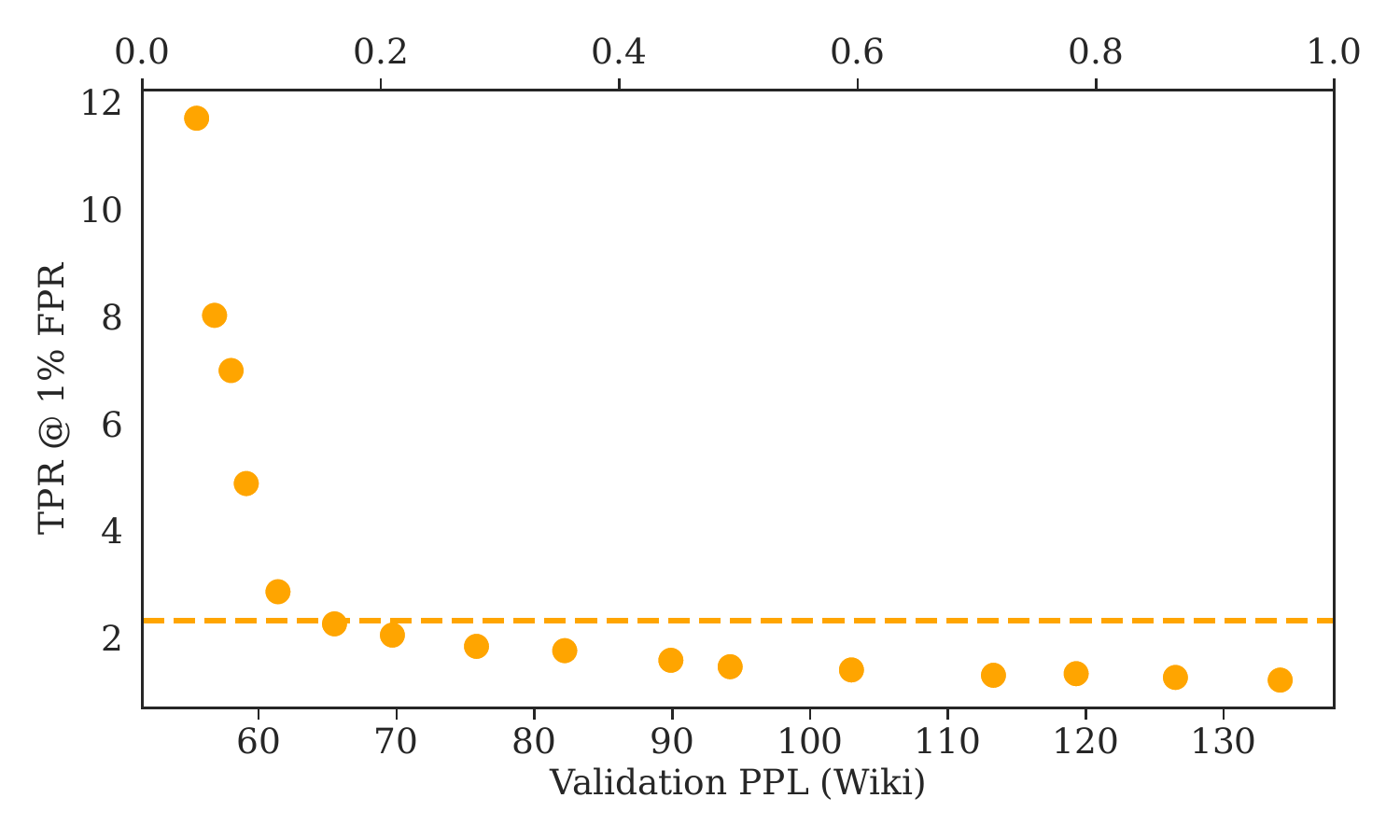}
\end{subfigure}
\begin{subfigure}{\linewidth}
   \includegraphics[width=1\linewidth]{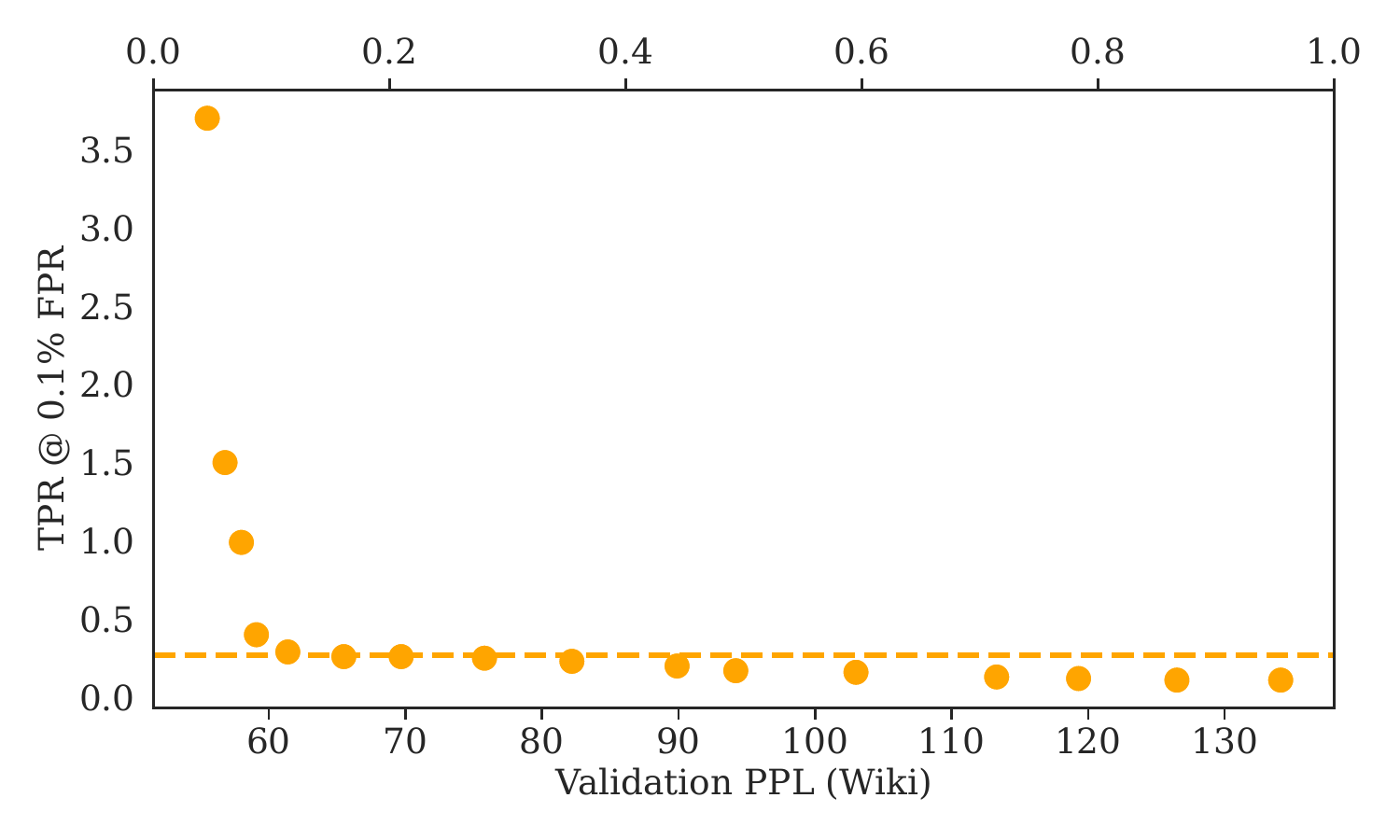}
\end{subfigure}
\begin{subfigure}{\linewidth}
   \includegraphics[width=1\linewidth]{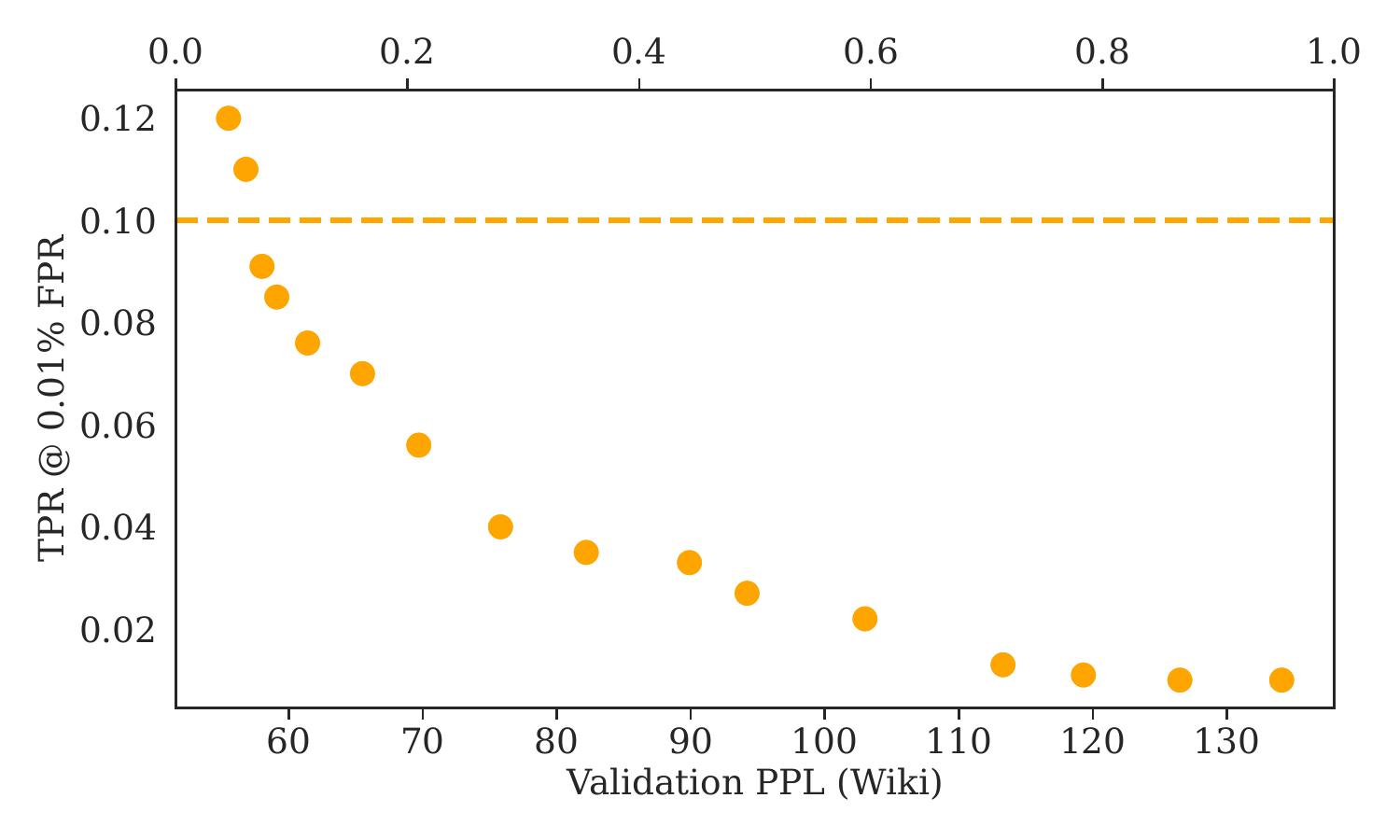}
\end{subfigure}
\caption{Attack Performance of reference attacks w.r.t validation PPL of reference models, compared to the performance of neighborhood attacks. The perplexity of the target model was 55.6 for Wikipedia}
\label{fig:epochstudywiki}

\end{figure}

\end{document}